\documentclass[twocolumn]{article}
\usepackage[margin=0.9in]{geometry}
\usepackage[utf8]{inputenc}
\usepackage[T1]{fontenc}
\usepackage{lmodern}
\usepackage{graphicx}
\usepackage{amsmath,amssymb}
\usepackage{hyperref}
\usepackage{booktabs}
\usepackage{array}
\usepackage{longtable}
\usepackage{tabularx}
\usepackage{caption}
\usepackage{float}
\usepackage{natbib}        
\usepackage{tikz}
\usetikzlibrary{arrows.meta,decorations.pathreplacing,positioning,shapes.geometric}
\usepackage{subcaption}

\title{MiCA Learns More Knowledge Than LoRA and Full Fine-Tuning}
\author{
  Sten R\"udiger\thanks{Independent Researcher, Email: \texttt{sten.ruediger@gmail.com}} 
  \and
  Sebastian Raschka\thanks{ RAIR Lab}
}

\begin{document}
\maketitle

\begin{abstract}
Minor Component Adaptation (MiCA) is a novel parameter-efficient fine-tuning method for large language models that focuses on adapting underutilized subspaces of model representations. Unlike conventional methods such as Low-Rank Adaptation (LoRA), which target dominant subspaces, MiCA leverages Singular Value Decomposition to identify subspaces related to minor singular vectors associated with the least significant singular values and constrains the update of parameters during fine-tuning to those directions. This strategy leads to up to 5.9x improvement in knowledge acquisition under optimized training hyperparameters and a minimal parameter footprint of 6-60\% compared to LoRA. These results suggest that constraining adaptation to minor singular directions provides a more efficient and stable mechanism for integrating new knowledge into pre-trained language models.

\end{abstract}

\section{Introduction}
The rapid advancement of large language models (LLMs) has introduced new challenges in adapting these models to specific tasks and domains without incurring the immense computational costs associated with full fine-tuning. Parameter-efficient fine-tuning (PeFT) methods have emerged as a viable alternative, allowing models to be adapted by updating a small subset of parameters while preserving the majority of the pre-trained model weights. Among the most prominent PeFT approaches is Low-Rank Adaptation (LoRA), which introduces trainable low-rank matrices into existing layers of the model to enable efficient adaptation \cite{hu2021lora}.
\begin{figure*}[t]
\centering

\begin{subfigure}[t]{0.49\textwidth}
\centering
\resizebox{0.9\linewidth}{!}{%
\begin{tikzpicture}[
  font=\sffamily,
  >=Latex,
  arr/.style={draw=blue!60!black, line width=1.2pt, -{Latex[length=3mm,width=2mm]}},
  brace/.style={decorate, decoration={brace, amplitude=4pt}, draw=blue!40!black, line width=1pt},
  boxW/.style={draw=black, fill=blue!40!white, minimum width=4.0cm, minimum height=4.0cm, align=center},
  bar/.style={draw=black, fill=yellow!45, minimum width=4.0cm, minimum height=0.35cm},
  trapTop/.style={draw=black, fill=orange!80, trapezium,
                  trapezium angle=110, minimum width=4.2cm, minimum height=1.4cm, align=center},
  trapBot/.style={draw=black, fill=orange!80, trapezium,
                  trapezium angle=70, minimum width=1.2cm, minimum height=1.4cm, align=center},
]

\newcommand{\customtrap}[7]{%
  \begin{scope}[shift={#2}]
    \pgfmathsetlengthmacro{\Tw}{#3}
    \pgfmathsetlengthmacro{\Bw}{#4}
    \pgfmathsetlengthmacro{\H}{#5}
    \path[draw=black, fill=#6]
      (-0.5*\Tw,  0.5*\H) --
      ( 0.5*\Tw,  0.5*\H) --
      ( 0.5*\Bw, -0.5*\H) --
      (-0.5*\Bw, -0.5*\H) -- cycle;
    \node[align=center] (#1) at (0,0) {#7};
  \end{scope}%
}

\node[boxW] (W) at (0,0) {%
  {\Large\bfseries Pre-trained}\\[2pt]
  {\Large\bfseries Weights}\\[12pt]
  {\Large $W \in \mathbb{R}^{d\times d}$}
};

\customtrap{B}{(4.3,1.3)}{4cm}{3cm}{1.4cm}{orange!80}{\Large $B=0$}

\customtrap{A}{(4.3,-1.3)}{3cm}{4cm}{1.4cm}{orange!80}{\Large $A=\mathcal{N}(0,\sigma^2)$}

\node[bar] (hbar) at (2.3,3.05) {};
\node[left=6pt of hbar, font=\Large\bfseries] (hlabel) {h};

\node[bar] (xbar) at (2.3,-3.35) {};
\node[left=6pt of xbar, font=\Large\bfseries] (xlabel) {x};

\node[font=\Huge, text=blue!60!black] (plus) at (2.1,2.4) {$+$};

\draw[arr] (0.4,2.15) -- (1.05,2.85);   
\draw[arr] (3.95,2.15) -- (3.4,2.85) ;   

\draw[arr] (1.05,-2.9) -- (0.35,-2.1) ; 
\draw[arr] (3.65,-2.9) -- (4.25,-2.1); 

\draw[brace] (2.8,-0.5) -- node[midway, yshift=10pt, font=\Large] {$r$} (5.8,-0.5);

\draw[brace] (0.3,-3.1) -- node[midway, yshift=12pt, font=\Large] {$d$} (4.3,-3.1);

\node[anchor=north west, font=\bfseries] at (current bounding box.north west) {(a)};
\end{tikzpicture}}
\end{subfigure}
\hfill
\begin{subfigure}[t]{0.49\textwidth}
\centering
\resizebox{0.9\linewidth}{!}{%
\begin{tikzpicture}[
  font=\sffamily,
  >=Latex,
  arr/.style={draw=blue!60!black, line width=1.2pt, -{Latex[length=3mm,width=2mm]}},
  brace/.style={decorate, decoration={brace, amplitude=4pt}, draw=blue!60!black, line width=1pt},
  boxW/.style={draw=black, fill=blue!40!white, minimum width=4cm, minimum height=4cm, align=center},
  bar/.style={draw=black, fill=yellow!45, minimum width=4.0cm, minimum height=0.35cm},
  trapTop/.style={draw=black, fill=orange!80, trapezium,
                  trapezium angle=110, minimum width=4.2cm, minimum height=1.4cm, align=center},
  trapBot/.style={draw=black, fill=orange!80, trapezium,
                  trapezium angle=70, minimum width=1.2cm, minimum height=1.4cm, align=center},
]

\newcommand{\customtrap}[7]{%
  \begin{scope}[shift={#2}]
    \pgfmathsetlengthmacro{\Tw}{#3}
    \pgfmathsetlengthmacro{\Bw}{#4}
    \pgfmathsetlengthmacro{\H}{#5}
    \path[draw=black, fill=#6]
      (-0.5*\Tw,  0.5*\H) --
      ( 0.5*\Tw,  0.5*\H) --
      ( 0.5*\Bw, -0.5*\H) --
      (-0.5*\Bw, -0.5*\H) -- cycle;
    \node[align=center] (#1) at (0,0) {#7};
  \end{scope}%
}

\node[boxW] (W) at (0,0) {%
  {\Large\bfseries Pre-trained}\\[2pt]
  {\Large\bfseries Weights}\\[12pt]
  {\Large $W \in \mathbb{R}^{d\times d}$}
};

\customtrap{B}{(4.3,1.3)}{4cm}{3cm}{1.4cm}{blue!40!white}{\Large $B=U_{[:,-r:]}$}

\customtrap{A}{(4.3,-1.3)}{3cm}{4cm}{1.4cm}{orange!80}{\Large $A=0$}

\node[bar] (hbar) at (2.3,3.05) {};
\node[left=6pt of hbar, font=\Large\bfseries] (hlabel) {h};

\node[bar] (xbar) at (2.3,-3.35) {};
\node[left=6pt of xbar, font=\Large\bfseries] (xlabel) {x};

\node[font=\Huge, text=blue!60!black] (plus) at (2.1,2.4) {$+$};

\draw[arr] (0.4,2.15) -- (1.05,2.85);   
\draw[arr] (3.95,2.15) -- (3.4,2.85) ;   

\draw[arr] (1.05,-2.9) -- (0.35,-2.1) ; 
\draw[arr] (3.65,-2.9) -- (4.25,-2.1); 

\draw[brace] (2.8,-0.5) -- node[midway, yshift=10pt, font=\Large] {$r$} (5.8,-0.5);

\draw[brace] (0.3,-3.1) -- node[midway, yshift=12pt, font=\Large] {$d$} (4.3,-3.1);

\node[anchor=north west, font=\bfseries] at (current bounding box.north west) {(b)};
\end{tikzpicture}}
\end{subfigure}

\caption{(a) Diagram of a LoRA-style low-rank adaptation module applied to a weight matrix. (b) Diagram of a MiCA-style adaptation module illustrating its distinct update structure compared to LoRA (blue: constrained components, orange: fine-tuned components initialized as indicated in each trapezoid). $U[:, -r:]$ means all rows, last $r$ columns of the matrix $U$ defined in the text.}
\label{fig:loramica}
\end{figure*}
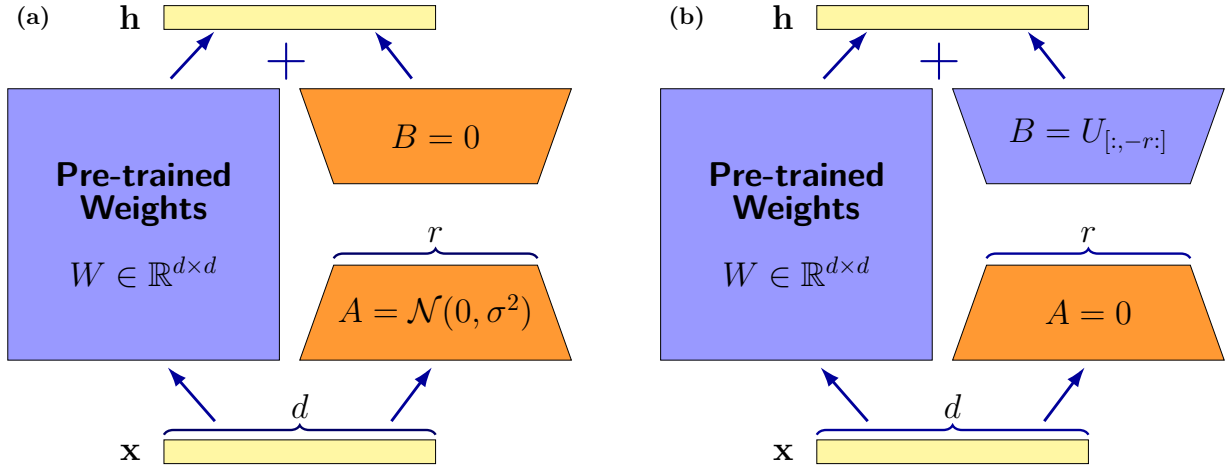





MiCA (Minor Component Adaptation) extends the capabilities of existing PeFT methods by introducing a novel strategy to identify and adapt the most informative and underutilized subspaces of model representations. MiCA aims to minimize parameter footprint, maximize knowledge acquisition, and reduce catastrophic forgetting.


To empirically validate the effectiveness of MiCA, a comprehensive experimental setup will be described, including the datasets, models, and comparative methods used in our evaluation. The results and analysis of these experiments will then be presented and discussed, highlighting the performance of MiCA across various tasks and benchmarks. Finally, we will touch upon potential applications and use cases for MiCA, followed by a concluding summary of our contributions and an outlook on future research directions.

\section{Related Work}
Parameter-efficient fine-tuning strategies have garnered attention as practical alternatives to full model fine-tuning, particularly in the era of large language models with billions of parameters \cite{lester2021power,li2021prefixtuning,benzaken2021bitfit}. These methods aim to preserve most of the pre-trained model while introducing lightweight, trainable components that adapt the model to new tasks. Among these, LoRA modules have been particularly influential.

\subsection{Low-Rank Adaptation}
LoRA inserts pairs of low-rank matrices into the existing weight matrices of a model. During training, only these low-rank matrices are updated, while the original weights remain frozen. The key insight behind LoRA is that model updates can often be effectively represented in a low-dimensional subspace. By decomposing the weight update into the product of two smaller matrices (typically much smaller than the original parameter space), LoRA significantly reduces the number of trainable parameters and memory footprint.

Let $W \in \mathbb{R}^{d \times d}$ denote a pre-trained weight matrix 
in a transformer layer (e.g., a projection matrix). 
Here, $d$ denotes the dimension of the layer, and $r \ll d$ 
is the rank of the update.

In parameter-efficient fine-tuning methods such as LoRA, 
the weight matrix is modified via a low-rank update:

\[
W_{\text{final}} = W + \Delta W,
\qquad
\text{where }
\Delta W = \frac{\alpha}{r}BA.
\]

The matrices are defined as:
\begin{itemize}
    \item $A \in \mathbb{R}^{r \times d}$,
    \item $B \in \mathbb{R}^{d \times r}$,
    \item $\alpha$ is a scaling parameter that determines how strongly the trained adapter weights influence the original pre-trained model weights, and
    \item $d$ is the dimension of $W$.
\end{itemize}

Since $r \ll d$, the update $\Delta W$ has rank at most $r$, 
allowing efficient adaptation with substantially fewer trainable parameters 
than full fine-tuning. During training, $W$ remains frozen and 
only $A$ and $B$ are optimized.

\medskip

Although we describe the square case for clarity of exposition, 
transformer weight matrices are in general rectangular, 
$W \in \mathbb{R}^{d_{\text{out}} \times d_{\text{in}}}$. 
The formulation extends directly by choosing 
$A \in \mathbb{R}^{r \times d_{\text{in}}}$ and 
$B \in \mathbb{R}^{d_{\text{out}} \times r}$, 
so that $\Delta W = BA$ matches the shape of $W$. 
All arguments presented in this work apply analogously to the general case.

Low-rank adaptation methods such as LoRA have become a standard 
approach for parameter-efficient fine-tuning of large language models. 
Instead of updating the full weight matrix $W$, only a rank-$r$ 
correction $\Delta W$ is learned, reducing the number of trainable 
parameters from $\mathcal{O}(d^2)$ to $\mathcal{O}(rd)$. 
This substantially lowers memory consumption, optimizer state size, 
and communication cost in distributed training.

Beyond computational efficiency, low-rank updates also act as an 
implicit regularizer. By constraining adaptation to a low-dimensional subspace, they reduce the risk of catastrophic forgetting and often  preserve pre-trained capabilities more effectively than full fine-tuning \cite{biderman2024lora}.

\subsection{Limitations of Current Approaches}

Low-rank adaptation methods such as LoRA have demonstrated strong empirical performance and substantial parameter efficiency. However, they leave a central question unresolved that is, \emph{which subspace should be adapted?}

In standard LoRA, both low-rank factors are optimized freely, meaning that the effective adaptation subspace is learned implicitly during training and may drift over time. While this flexibility can be beneficial, it does not explicitly account for the spectral structure of the pre-trained weight matrix. In particular, conventional low-rank adaptation does not distinguish between dominant and minor directions in weight space.

As a consequence, updates may overlap with high-energy components that encode general pre-trained capabilities. Such overlap can lead to interference, inefficient use of representational capacity, or undesirable shifts in previously acquired knowledge. Although parameter-efficient fine-tuning methods are generally less prone to catastrophic forgetting than full fine-tuning \cite{chen2020continual,ramasesh2022effect}, they do not explicitly optimize for knowledge retention, and degradation on unrelated tasks can still occur.

Several recent approaches incorporate singular value decomposition (SVD) into the adaptation process, such as PiSSA \cite{meng2024pissa}. However, these methods primarily use spectral information for initialization; the adaptation parameters are subsequently free to evolve during training \cite{yun2025soma,lee2026learning}. Consequently, the effective subspace may again drift toward dominant components.

These observations suggest that low-rank adaptation alone is not sufficient and that the \emph{choice and stability of the adaptation subspace} may play a crucial role. MiCA addresses this limitation by explicitly constraining learning to fixed minor singular directions, introducing a spectrally grounded mechanism for parameter-efficient specialization.

The core hypothesis of MiCA is therefore that a subset of minor components, those associated with low singular values in a weight matrix, can be activated for critical task-specific adaptability. These components are typically discarded or ignored in conventional low-rank approximations, which prioritize dominant directions. MiCA inverts this logic and proposes that the less-expressed directions, while accounting for a smaller portion of the original matrix weights, may offer higher marginal utility for task-specific learning. It holds especially when the dominant subspace is already saturated with generic, pre-trained knowledge. This leads to a refinement of the PeFT principle: rather than merely reducing dimensionality, we should aim to adapt the model in directions that are both underused and maximally informative for the new task.

MiCA bears similarity to Minor Component Analysis (MCA) \cite{oja1992principal}, a lesser-known counterpart to Principal Component Analysis (PCA). While PCA extracts dominant eigenvectors associated with the largest eigenvalues (capturing maximum variance), MCA focuses on eigenvectors associated with the smallest eigenvalues. These minor components are typically orthogonal to the dominant subspace and may contain weak signals or structure that PCA overlooks. 
Analogously, MiCA leverages MCA-like logic to identify hidden components in the weight space directions that are less significant in a variance sense but potentially more adaptable, plastic, or task-relevant.

By adapting these minor components instead of reusing or reinitializing arbitrary low-rank directions, MiCA offers a data-driven, semantically grounded mechanism for efficient model adaptation thus preserving core model capabilities while enhancing domain-specific expressiveness.

\section{Methodology}

MiCA extends LoRA’s architecture by explicitly incorporating the SVD of the pre-trained weight matrix.

Let $W \in \mathbb{R}^{d \times d}$ denote a pre-trained weight matrix in a transformer layer. 
Its singular value decomposition is given by

\[
W = U \Sigma V^\top,
\]

where:
\begin{itemize}
    \item $U \in \mathbb{R}^{d \times d}$ is an orthogonal matrix whose columns are the left singular vectors,
    \item $V \in \mathbb{R}^{d \times d}$ is an orthogonal matrix whose columns are the right singular vectors,
    \item $\Sigma \in \mathbb{R}^{d \times d}$ is a diagonal matrix containing the singular values $\sigma_1 \geq \sigma_2 \geq \dots \geq \sigma_d \geq 0$.
\end{itemize}

Although we describe the square case for clarity, the formulation extends directly to rectangular weight matrices $W \in \mathbb{R}^{d_{\text{out}} \times d_{\text{in}}}$ via the standard SVD.

\subsection{Integration with Transformer Layers}

MiCA integrates into attention layers by adding low-rank matrices 
$A \in \mathbb{R}^{r \times d}$ and $B \in \mathbb{R}^{d \times r}$, 
initialized based on the singular vectors derived from SVD. The weight update is defined as before:

\[
\Delta W = \frac{\alpha}{r} BA
\]

The final adapted weights in the transformer layers are computed as before in LoRA:

\[
W_{\text{final}} = W + \Delta W.
\]

Following the original LoRA study, adapters were applied to the query and value projection matrices ($q_{\text{proj}}$, $v_{\text{proj}}$) in each transformer block unless otherwise stated.

\subsection{Parameter Selection and Injection Points}
The rank $r$ is chosen based on computational constraints and the desired balance between adaptability and computational efficiency. 
MiCA specifically selects the least significant singular vectors from the decomposition to be updated with new knowledge. For a constrained output space:

\begin{itemize}
        \item Compute the complete SVD: $W = U \Sigma V^T$
        \item Select the last  $r$ singular vectors of: $U_r = U[:, -r :]$
        \item Initialize: $B = U_r, \; A = 0$
\end{itemize}

Note that with this choice, the updated matrix projects into subspaces that belong to small singular values. 
This targeted initialization helps to keep subspaces injecting parameters efficiently at critical points within transformer layers, typically at attention.

\subsection{Training Procedure}
\label{sec:trainingproc}
During training, MiCA adapts transformer layers by optimizing the matrix $A$ via gradient descent 
while keeping the original weights $W$ and the matrix $B$ frozen. The objective function is 
standard for transformer-based fine-tuning tasks, with an added low-rank update structure.

We use a fine-tuning strategy that leverages the separation of pre-training and instruction-tuning 
phases commonly used in large language models \cite{wei2021finetuned,ouyang2022training,dettmers2023qlora}. Rather than fine-tuning an 
instruction-tuned model directly, we fine-tune the \textbf{base model} 
(i.e., the model pre-trained without instruction-following capabilities) on the target task. 
After this task-specific fine-tuning, we \textit{additively compose} the resulting model weights 
with the delta between the instruction-tuned model and its corresponding base model:

\[
\theta_{\text{final}} = \theta_{\text{base}}^{\text{FT}} + \left( \theta_{\text{instr}} - \theta_{\text{base}} \right)
\]

Here, $\theta_{\text{base}}^{\text{FT}}$ represents the fine-tuned base model, and 
$\left( \theta_{\text{instr}} - \theta_{\text{base}} \right)$ denotes the instruction delta 
derived from the original instruction tuning. This technique assumes a near-linear locality in the 
weight space and builds on the empirical observation that instruction tuning modifies a relatively 
constrained set of parameters relevant to instruction adherence.









This approach offers two key advantages: 
(1) it enables the model to retain strong task-specific alignment learned during fine-tuning 
on domain-specific data, and 
(2) it reintroduces general instruction-following capabilities without requiring costly 
end-to-end instruction re-tuning. 
Preliminary tests suggested improved composability and reduced interference compared to 
fine-tuning on top of instruction-tuned checkpoints directly.

\section{Experimental Setup}
To assess the effectiveness of MiCA in improving fine-tuning quality, we conducted a suite of experiments across various datasets, models, and methods. Our experiments address multiple dimensions: knowledge retention, domain generalization, catastrophic forgetting, and parameter efficiency.

We explored several training and testing setups that span general language understanding, synthetic question answering, coding tasks, and domain-specific applications:

\begin{itemize}
\item OpenAI blog posts from May and June 2024 (called BLOGS dataset below) together with related synthetic QAs for multiple-choice tasks for domain question answering performance (BLOGS-MC). The posts were published after the knowledge cutoff of the LLMs tested below.
\item TruthfulQA \cite {lin2021truthfulqa} and Hellaswag \cite{zellers2019hellaswag} benchmarks for adversarial robustness and factuality.
\item A German history book \cite{ruediger2024astronomen} from 2024 that is similarly not in training sets of foundational models from before 2024.
\item Testing data include target domains (to evaluate domain-specific performance) and source domains (to measure knowledge retention and catastrophic forgetting).
\end{itemize}

To test the performance of MiCA, we used:
\begin{itemize}
\item Full fine-tuning (Full FT): updating the entire model to learn domain-specific knowledge,
\item LoRA,
\item Pre-trained models (Llama-2-7B, \cite{touvron2023llama2} with a knowledge cutoff July 2023 and Qwen-2.5 \cite{qwen2025qwen25technicalreport} with a knowledge cutoff October 2023) as non-adapted references.
\end{itemize}

This experimental framework provides our basis for evaluating MiCA's impact on fine-tuning outcomes across different scenarios, target domains, and parameter constraints. The technical details of our method can be found in the appendix.

\section{Results and Analysis}
This section presents the findings from our experimental evaluation of MiCA compared to other fine-tuning methods, particularly LoRA and full fine-tuning. Our experiments span multiple training–testing setups, including continued pre-training on texts of different sizes, on synthetic multiple-choice tasks, and general knowledge benchmarks.  These controlled cross-domain evaluations allow us to assess MiCA’s improvements in knowledge acquisition, its robustness when transferring from heterogeneous training sources to  evaluation domains, its mitigation of catastrophic forgetting, and its parameter efficiency compared to LoRA.

\subsection{BLOGS dataset}

To evaluate the knowledge uptake and retention capabilities of MiCA, we first conducted continued pre-training experiments on the Llama-2-7B and Qwen2.5-7B models. The training corpus, BLOGS, consisted of ten  blog articles from OpenAI's website (300 to 600 words per post), each expanded via GPT-4-generated paraphrasing to produce a total of thirty texts. These documents introduced previously unseen domain knowledge that is not contained in the base model, allowing a controlled study of how effectively MiCA incorporates novel information.

To quantify knowledge uptake, we constructed a synthetic evaluation dataset, BLOGS-MC,  comprising 300 multiple-choice questions derived directly from blog content by asking GPT-4 \cite{openai2024gpt4technicalreport}. 
We refer to the resulting evaluation benchmark as BLOGS-MC. For broader generalization effects, we additionally evaluated  the TruthfulQA and Hellaswag benchmarks.

Both MiCA and LoRA were trained under an independently optimized hyperparameter search. The search space included adapter rank $r$, learning rate, and number of training epochs. To ensure a fair comparison, all experiments used identical training scripts, tokenization settings, and data orderings, each configuration being evaluated across multiple random seeds. 

\begin{figure}[h!]
    \centering

    \begin{subfigure}[t]{0.9\linewidth}
        \centering
        \includegraphics[width=\linewidth]{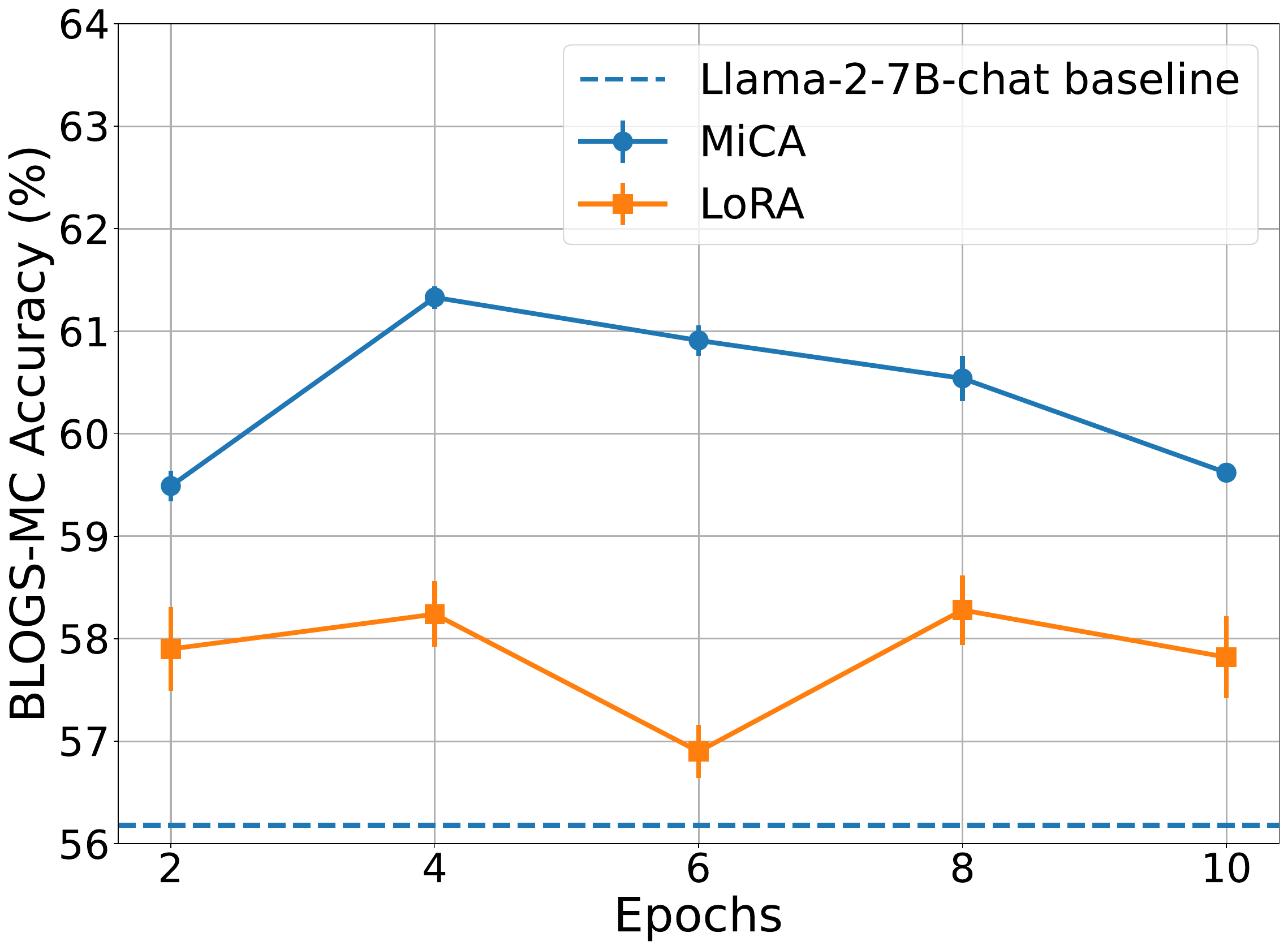}
        \caption{Baseline: Llama-2-7B}
        \label{fig:accuracy_epochs_a}
    \end{subfigure}

    \vspace{0.5em}

    \begin{subfigure}[t]{0.9\linewidth}
        \centering
        \includegraphics[width=\linewidth]{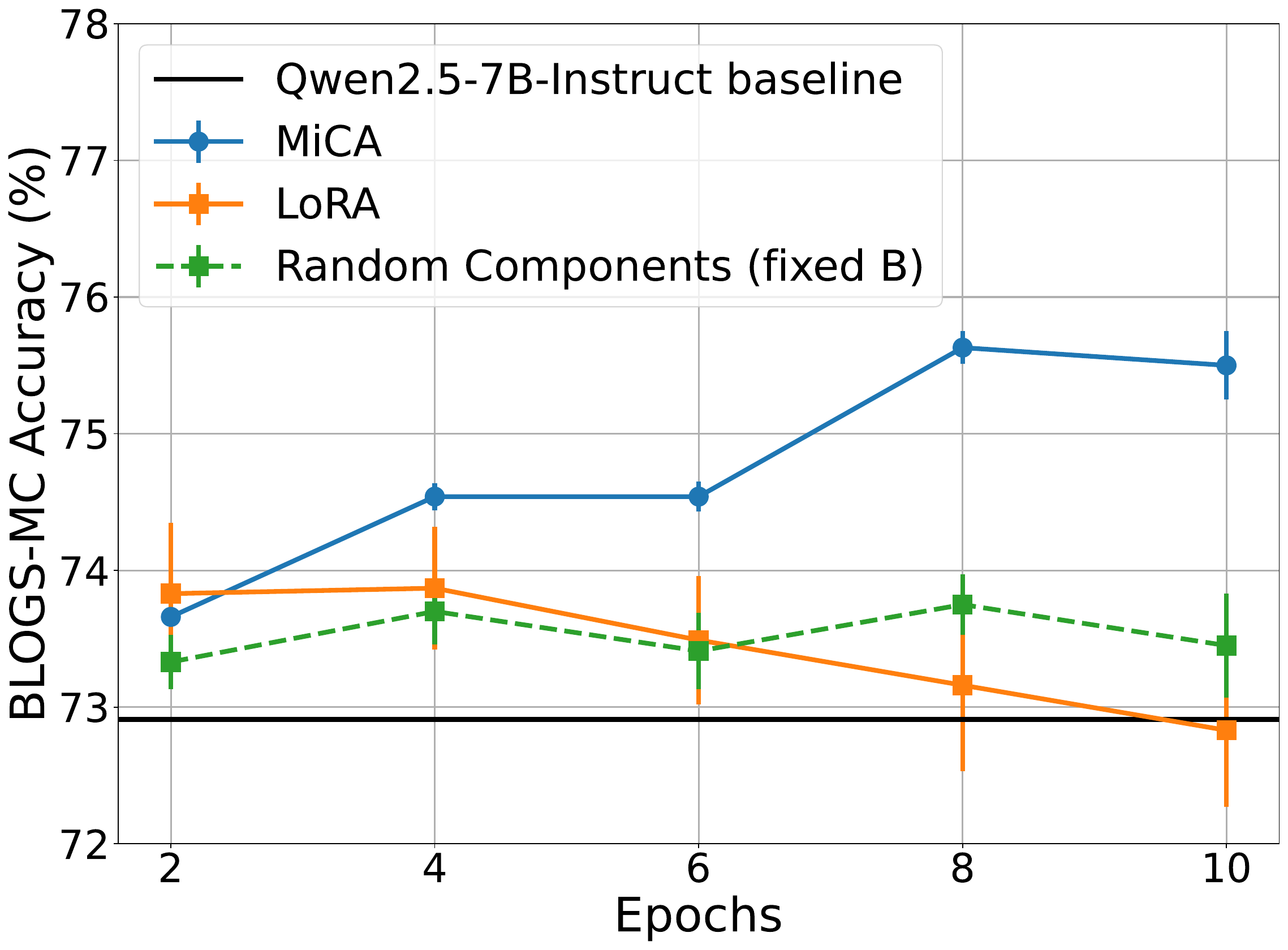}
        \caption{Baseline: Qwen2.5-7B}
        \label{fig:accuracy_epochs_b}
    \end{subfigure}

    \caption{Retention of domain knowledge from the BLOGS dataset as a function of number of training epochs for MiCA and LoRA. Error bars denote the standard error of 8 evaluation runs. The horizontal dashed line shows the baseline performance of the respective foundation model. The green dashed line in (b) shows results for a fixed $B$ matrix initialized with random SVD components instead of the minor componenents and same hyperparameters as for the MiCA runs.}
    \label{fig:accuracy_epochs}
\end{figure}

\begin{table*}[th!]
\centering
\begin{tabular}{lrrrrrrr}
\hline
\textbf{Method} & \textbf{BLOGS-MC} & \textbf{TruthQA} & \textbf{HellaSwag} & \textbf{r} & \textbf{LR} & \textbf{Ep.} & \textbf{Params}\\
\hline
Llama-2-7B-chat & 56.18    & 34.79  & \bf{60.40} & -- & -- & -- & 6,747M \\
LoRA (optimal)& 58.28 & {\bf 35.47} & 60.41 & 128 & 1e-4 & 8 & 67M\\
MiCA (optimal)  & \bf{61.33}    & 35.29 & 60.11  & 16 & 5e-4 & 4 & 4M \\
\hline
Qwen2.5-7B-Instr. & 72.91   & 43.27      &  60.60       & -- & -- & -- & 7,626M\\
LoRA (optimal)      & 73.87   & 42.95      &  60.95      & 32 & 5e-4 & 4 & 10M\\
MiCA  (optimal) &  \bf{75.63} & \bf{43.38} &  \bf{61.62} & 32 & 5e-4 & 8 & 6M\\
\hline
\end{tabular}
\caption{Comparison of BLOGS-MC (new knowledge), TruthfulQA-MC1 and HellaSwag accuracies across different fine-tuning methods for the blog post training. For each accuracy value we averaged 8 runs per hyperparameter choice. Additional columns for optimal adapter rank ($r$), learning rate (LR), and training epochs are included for completeness. The last columns shows the number of parameters of the full model or the number of  trainable  parameters for the LoRA and MiCA runs, respectively. }
\label{tab:truthfulqa-updated}
\end{table*}

Figure~\ref{fig:accuracy_epochs} shows the accuracy of BLOGS-MC as a function of the number of continued pre-training epochs for both the Llama (a) and Qwen models (b). Each curve represents the mean performance of eight independent training runs with different random seeds, while the error bars denote the corresponding standard error. 
Across both models, MiCA consistently achieves higher domain-knowledge accuracy than LoRA throughout training. The improvement emerges already in the early epochs and remains stable as training progresses, suggesting that restricting updates to minor singular directions enables more efficient knowledge integration.

Table~\ref{tab:truthfulqa-updated} summarizes the performance of MiCA and LoRA on the 300-item multiple-choice knowledge test, alongside their respective levels of general knowledge and reasoning as measured by TruthfulQA (mc1) and Hellaswag. It reports the best-performing configuration from the training curves shown in Figure~\ref{fig:accuracy_epochs}. Across both model families, MiCA achieves the highest BLOGS-MC accuracy. For Llama-2-7B-chat, MiCA reaches 61.33\%, improving over the baseline model (56.18\%) by 5.15 percentage points and over the best LoRA configuration (58.28\%) by 3.05 points. A similar pattern is observed for Qwen2.5-7B-Instruct, where MiCA achieves 75.63\% accuracy compared to 73.87\% for LoRA and 72.91\% for the instruct baseline.



MiCA achieves the highest BLOGS-MC accuracy while using substantially fewer trainable parameters than LoRA. For the Llama model, MiCA uses an adapter rank of r = 16, while LoRA requires r = 128 to reach its best precision. Moreover, MiCA updates only one of the two LoRA matrices, effectively halving the parameter count relative to standard LoRA for the same rank. Taken together, this results in about 6\% trainable parameters for the MiCA configuration, achieving competitive or superior performance compated to LoRA. For the Qwen model, both methods reach highest BLOG score for $r=32$, but because of the freezing of $B$, MiCA still needs only about 60\% of the parameter count.



Overall, the results demonstrate that MiCA provides a more parameter-efficient and representation-stable fine-tuning strategy for acquiring new knowledge in Llama-2-7B-chat, delivering higher accuracy and lower computational footprint compared to LoRA.
However, these results alone do not establish whether the improvement arises specifically from adapting minor singular directions or from restricting updates to any fixed low-rank subspace. We investigate this question in the ablation study in Section 7.

\subsection{HISTORY book in German}

\begin{table*}[h!]
\centering
\begin{tabular}{lccccc}
\toprule
\textbf{Method} & \textbf{Accuracy (\%)} & \textbf{HellaSwag (\%)} & \textbf{Learning rate} & \textbf{r} & \textbf{\# epochs} \\
\midrule
Llama-2-7B-chat        & 27.4    & \bf{57.8} &   --  & --      & -- \\
Full FT (chat)         & 30.4    & 57.3 &  2e-5   & --      & 1  \\
LoRA       & 29.4\footnotemark[1] & 57.7   & 5e-4   & 32      & 8  \\
MiCA                   & \bf{39.2}\footnotemark[2]  & \bf{57.8} & 2e-3 & 32      & 8  \\
\bottomrule
\end{tabular}
\caption{Retention of new knowledge for the HISTORY dataset: accuracy on multiple choice tests for several fine-tuning methods (Llama-2) with respective optimal hyperparameter choices. ($^1$ mean of results for two runs, $^2$ identical results across four independent runs)}
\end{table*}

A second continued pre-training test with a much longer text base involved a 300-page history book in German (HISTORY dataset, approximately 100,000 tokens). The evaluation of the model's ability to acquire new knowledge was performed using a custom-built dataset HISTORY-MC of 102 multiple-choice questions (A-D) based on the book's content (questions and answers were generated with Claude Sonnet 3.5). The base model Llama-2-7B was used for training, and the resulting PeFT matrices were subsequently merged with the instruct variant of the model as described in Section \ref{sec:trainingproc}.

\begin{figure}[]
    \centering
    \includegraphics[width=1.0\linewidth]{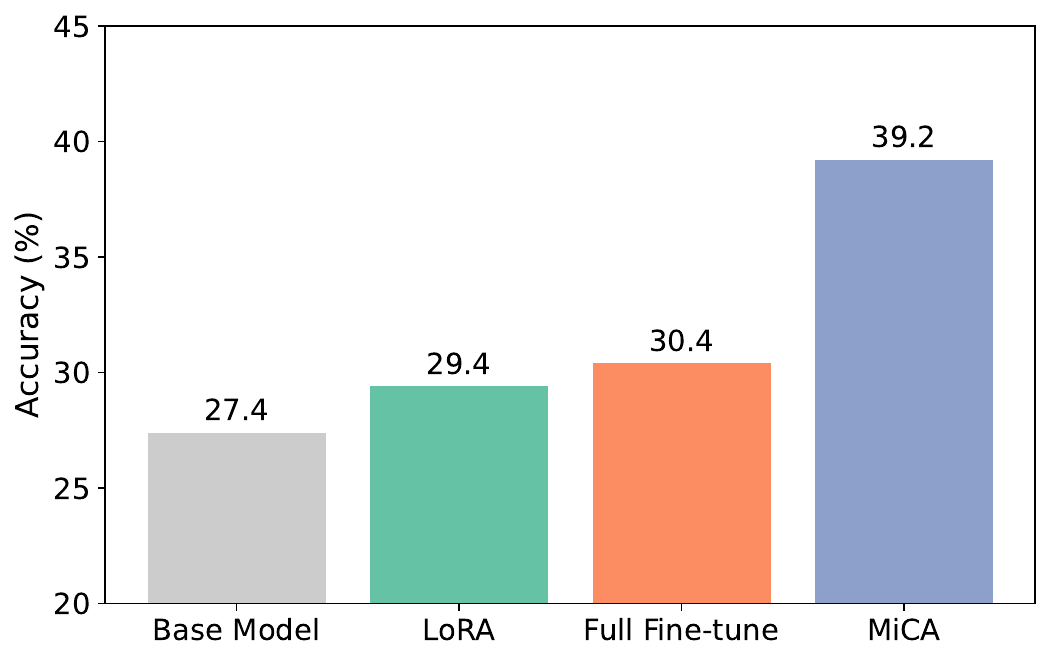}
    \caption{Retention of domain knowledge for the history book and for several fine-tuning methods for Llama-2. }
    \label{fig:accuracy_book}
\end{figure}

As before, hyper-parameters learning rate, number of epochs (with a maximum of 8), and rank ($r$) were optimized for both LoRA and MiCA. The results (see Figure~\ref{fig:accuracy_book}) indicate that MiCA, with optimal parameters, achieved a significantly higher precision in answering questions about new knowledge (39.2\% with four equivalent results) compared to LoRA (29.4\% on average over two runs) and the fully fine-tuned chat model (30.4\%), which were also independently optimized for hyperparameters. The original chat variant of the model achieved an accuracy of 27.4\%. Analysis of the HellaSwag dataset yielded similar results for PeFT methods (approximately 0.57-0.58 Acc) but significantly mode forgetting for the full fine-tuning method. The optimal learning rates for MiCA and LoRA differed (2e-3 vs. 5e-4), while both methods achieved the best results with a rank of 32 and 8 training epochs. The fully fine-tuned model used a learning rate of 2e-5 and was trained for only one epoch.

The observation of minimal catastrophic forgetting is potentially explained by the fact that training was conducted in German, while general knowledge tests were performed in English (HellaSwag). 

As in the case of BLOGS, MiCA is more parameter-efficient and achieved the most significant improvement in accuracy on new knowledge with about half of the number of trainable parameters as LoRA.
Comparing the small-scale blog experiments with the substantially larger history corpus suggests an intriguing pattern, the performance gap between MiCA and LoRA widens as domain data increases. This suggests that constraining adaptation to minor singular directions becomes increasingly beneficial in higher-signal regimes, where unconstrained updates risk overwriting dominant pre-trained structure. The results further indicate that spectrally localized adaptation mitigates interference even under extended continued pre-training.

\section{Ablation: Minor vs.\ Major Singular Subspaces}

To isolate whether MiCA's gains arise specifically from adapting \emph{minor} singular directions rather than from any fixed orthonormal subspace, we compare four variants on Qwen-2.5-7B training on the blog posts:

\begin{enumerate}
    \item \textbf{Instruct baseline} (no fine-tuning)
    \item \textbf{Major-$r$ adaptation}: projection onto the top-$r$ singular vectors
    \item \textbf{Minor-$r$ adaptation (MiCA)}: projection onto the bottom-$r$ singular vectors
        \item \textbf{Random components}: the matrix $B$ is fixed and initialized from $r$ randomly chosen components of the SVD
\end{enumerate}

The random-component variant serves as a control to test whether MiCA’s gains arise merely from restricting adaptation to a fixed low-dimensional orthonormal subspace. If this were the case, randomly selected singular vectors should perform comparably to minor components. 

All three adapted variants use the same rank $r$, learning rate, number of steps, and trainable parameter count. In all cases, the projection matrix $B$ is fixed and only the coefficient matrix $A$ is trained. Thus, the only varying factor is the choice of singular subspace.

\begin{table}[h]
\centering
\begin{tabular}{lcc}
\toprule
Method & BLOGS-MC (\%)  \\
\midrule
Instruct (no FT) & 72.91  \\
Major-$r$ & 74.21  \\
$r$ random components & 73.75  \\
\textbf{Minor-$r$ (MiCA)} & \textbf{75.63}  \\
\bottomrule
\end{tabular}
\caption{Ablation on Qwen-2.5-7B comparing adaptation in major vs.\ minor and randomly chosen singular subspaces.}
\label{tab:minor_major_ablation}
\end{table}

All subspace-restricted adaptations improve over the instruct baseline on the domain BLOGS-MC benchmark, indicating that constrained low-rank adaptation alone provides benefits. However, adapting along \emph{minor} singular directions outperforms major-direction and random component adaptation, yielding improvement over both other choices.

Fig.~\ref{fig:accuracy_epochs} B also contains the random-component results across training epochs. While random directions initially follow a similar improvement trajectory as the other subspace-restricted methods, they plateau earlier and converge to a lower final accuracy than MiCA. This suggests that the minor singular subspace not only provides higher final performance but also supports more stable adaptation dynamics during training.



Taken together, the results provide direct empirical support for MiCA’s central hypothesis: minor singular directions form a more plastic subspace for domain adaptation, while dominant or random directions are structurally less plastic.

\section{Discussion, Limitations, and Outlook}

MiCA introduces a spectrally grounded approach to parameter-efficient fine-tuning by constraining adaptation to minor singular directions of pre-trained weights. Empirically, this yields improved knowledge acquisition, potentially reduced catastrophic forgetting, and significantly lower parameter counts compared to LoRA under matched optimization.

\subsection{Limitations}

MiCA in the current setup is not designed for instruction fine-tuning. Since it restricts adaptation to minor spectral components of the base model, tasks that primarily require structural instruction-following rather than knowledge integration may not benefit from this constraint. 
However, future research may explore the potential of applying Singular Value Decomposition to the delta weights obtained from instruction fine-tuning processes versus the base model. This could potentially pave the way for extending the benefits of MiCA to instruction-based tasks by identifying and adapting only the most salient weight changes.

Additionally, while we demonstrate consistent improvements on 7B-scale models, the scaling behavior with respect to model size, dataset size, and task complexity remains an open question. Finally, computing SVDs for very large layers may introduce additional preprocessing cost, although this is a one-time operation.


\subsection{Practical Implications}
MiCA is particularly suited for domain specialization scenarios where integrating new factual knowledge while preserving pre-trained capabilities is critical. Its reduced parameter footprint makes it attractive for on-device adaptation and federated learning settings, where communication and memory constraints are central. These properties suggest that spectrally constrained adaptation may serve as a principled foundation for deploying lightweight, domain-specific LLM variants.

An interesting extension is to combine spectrally constrained adaptation with reinforcement learning objectives. Since MiCA restricts updates to minor singular directions, it localizes representational change. Reinforcement learning with explicit KL regularization has been observed to produce more behaviorally targeted updates and, in some settings, reduced catastrophic forgetting compared to supervised fine-tuning. A two-stage pipeline — (1) continued pre-training with MiCA for knowledge injection, followed by (2) RL-based alignment on labeled or preference data — may further stabilize domain adaptation while preserving pre-trained capabilities. Investigating whether spectral constraints and reward-based optimization act synergistically in reducing interference is a promising direction for future work.





\bibliography{cite}

@article{hu2021lora,
  author  = {Hu, Edward J. and Shen, Yelong and Wallis, Phillip and Allen-Zhu, Zeyuan and Li, Yuanzhi and Wang, Shean and Wang, Lu and Chen, Weizhu},
  title   = {LoRA: Low-Rank Adaptation of Large Language Models},
  journal = {arXiv preprint arXiv:2106.09685},
  year    = {2021},
  url     = {https://arxiv.org/abs/2106.09685}
}

@article{lester2021power,
  author  = {Lester, Brian and Al-Rfou, Rami and Constant, Noah},
  title   = {The Power of Scale for Parameter-Efficient Prompt Tuning},
  journal = {arXiv preprint arXiv:2104.08691},
  year    = {2021},
  url     = {https://arxiv.org/abs/2104.08691}
}

@inproceedings{li2021prefixtuning,
  title        = {Prefix-Tuning: Optimizing Continuous Prompts for Generation},
  author       = {Li, Xiang Lisa and Liang, Percy},
  booktitle    = {Proceedings of ACL 2021},
  pages        = {4582--4597},
  year         = {2021},
  url          = {https://arxiv.org/abs/2101.00190}
}

@inproceedings{benzaken2021bitfit,
  title        = {BitFit: Simple Parameter-efficient Fine-tuning for Transformer-based Masked Language-models},
  author       = {Ben-Zaken, Elena and Ravfogel, Shauli and Goldberg, Yoav},
  booktitle    = {Proceedings of the 2021 Conference on Empirical Methods in Natural Language Processing},
  pages        = {1--9},
  year         = {2021},
  url          = {https://arxiv.org/abs/2106.10199}
}

@article{biderman2024lora,
  title={Lora learns less and forgets less},
  author={Biderman, Dan and Portes, Jacob and Ortiz, Jose Javier Gonzalez and Paul, Mansheej and Greengard, Philip and Jennings, Connor and King, Daniel and Havens, Sam and Chiley, Vitaliy and Frankle, Jonathan and others},
  journal={arXiv preprint arXiv:2405.09673},
  year={2024}
}

@article{oja1992principal,
  title        = {Principal and Minor Components in Neural Networks},
  author       = {Oja, Erkki},
  journal      = {Neural Networks},
  volume       = {5},
  number       = {6},
  pages        = {927--935},
  year         = {1992},
  publisher    = {Elsevier}
}

@article{wei2021finetuned,
  title        = {Finetuned Language Models are Zero-Shot Learners},
  author       = {Wei, Jason and Bosma, Maarten and others},
  journal      = {arXiv preprint arXiv:2109.01652},
  year         = {2021},
  url          = {https://arxiv.org/abs/2109.01652}
}

@inproceedings{ouyang2022training,
  title        = {Training Language Models to Follow Instructions with Human Feedback},
  author       = {Ouyang, Long and Wu, Jeff and Jiang, Xu and others},
  booktitle    = {Advances in Neural Information Processing Systems},
  year         = {2022},
  url          = {https://arxiv.org/abs/2203.02155}
}

@article{dettmers2023qlora,
  title        = {QLoRA: Efficient Finetuning of Quantized LLMs},
  author       = {Dettmers, Tim and Lewis, Mike and Shleifer, Sam and Zettlemoyer, Luke},
  journal      = {arXiv preprint arXiv:2305.14314},
  year         = {2023},
  url          = {https://arxiv.org/abs/2305.14314}
}

@article{lin2021truthfulqa,
  title        = {TruthfulQA: Measuring How Models Mimic Human Falsehoods},
  author       = {Lin, Stephanie and Hilton, Jacob and Evans, Owain},
  journal      = {Proceedings of ACL 2022},
  year         = {2021},
  url          = {https://arxiv.org/abs/2109.07958}
}

@inproceedings{zellers2019hellaswag,
  title        = {HellaSwag: Can a Machine Really Finish Your Sentence?},
  author       = {Zellers, Rowan and Holtzman, Ari and Bisk, Yonatan and Farhadi, Ali and Choi, Yejin},
  booktitle    = {Proceedings of ACL 2019},
  pages        = {4791--4800},
  year         = {2019},
  url          = {https://arxiv.org/abs/1905.07830}
}

@article{touvron2023llama2,
  title        = {LLaMA 2: Open Foundation and Fine-Tuned Chat Models},
  author       = {Touvron, Hugo and Martin, Louis and Stone, Kevin and others},
  journal      = {arXiv preprint arXiv:2307.09288},
  year         = {2023},
  url          = {https://arxiv.org/abs/2307.09288}
}

@inproceedings{chen2020continual,
  title     = {Continual Learning with Pretrained Language Models},
  author    = {Chen, Xudong and Shu, Lei and Ma, Tengfei and Ye, Xinyu and Carlini, Nicholas and Chen, Pin-Yu and Wang, Shiqi and Hsieh, Cho-Jui},
  booktitle = {Findings of the Association for Computational Linguistics: EMNLP 2020},
  pages     = {270--282},
  year      = {2020},
  url       = {https://arxiv.org/abs/2007.14413}
}

@article{ramasesh2022effect,
  title     = {Anatomy of Catastrophic Forgetting: Hidden Representations and Task Semantics},
  author    = {Ramasesh, Vinay V. and Dyer, Ethan and Raghu, Maithra},
  journal   = {Proceedings of ICLR 2022},
  year      = {2022},
  url       = {https://arxiv.org/abs/2007.07038}
}

@article{meng2024pissa,
  title={Pissa: Principal singular values and singular vectors adaptation of large language models},
  author={Meng, Fanxu and Wang, Zhaohui and Zhang, Muhan},
  journal={Advances in Neural Information Processing Systems},
  volume={37},
  pages={121038--121072},
  year={2024}
}

@article{ruediger2024astronomen,
  title={Astronomen, {A}kten und {A}ff{\"a}ren. {V}om {A}nfang zum {E}nde des {A}strophysik-alischen {O}bservatoriums zu {P}otsdam. {B}erichte und {E}rinnerungen},
  author={Rüdiger, Günther},
  journal={Journal of Astronomical History and Heritage},
  volume={27},
  number={3},
  pages={727--728},
  year={2024}
}

@inproceedings{yun2025soma,
  title={SoMA: Singular Value Decomposed Minor Components Adaptation for Domain Generalizable Representation Learning},
  author={Yun, Seokju and Chae, Seunghye and Lee, Dongheon and Ro, Youngmin},
  booktitle={Proceedings of the Computer Vision and Pattern Recognition Conference},
  pages={25602--25612},
  year={2025}
}

@article{lee2026learning,
  title={Learning Rate Matters: Vanilla LoRA May Suffice for LLM Fine-tuning},
  author={Lee, Yu-Ang and Ko, Ching-Yun and Chen, Pin-Yu and Yeh, Mi-Yen},
  journal={arXiv preprint arXiv:2602.04998},
  year={2026}
}

@misc{qwen2025qwen25technicalreport,
      title={Qwen2.5 Technical Report}, 
      author={Qwen},
      year={2025},
      eprint={2412.15115},
      archivePrefix={arXiv},
      primaryClass={cs.CL},
      url={https://arxiv.org/abs/2412.15115}, 
}

@misc{openai2024gpt4technicalreport,
      title={GPT-4 Technical Report}, 
      author={OpenAI},
      year={2024},
      eprint={2303.08774},
      archivePrefix={arXiv},
      primaryClass={cs.CL},
      url={https://arxiv.org/abs/2303.08774}, 
}


\onecolumn
\appendix
\section{Technical Details - BLOGS corpus}
\label{app:training-details}

This appendix documents the implementation details for Continued Pre-Training (CPT) with parameter-efficient adapters (LoRA / MiCA-style variants), including model choices, hyperparameters, prompting, data construction, and evaluation protocol. Unless stated otherwise, all experiments were run with fixed hyperparameters across methods to ensure comparability.

\subsection{Software Stack and Core Libraries}
\label{app:stack}
We used the Hugging Face \texttt{transformers} ecosystem together with TRL's SFT trainer:
\begin{itemize}
  \item \texttt{transformers} (model loading, tokenization, generation pipelines)
  \item \texttt{peft} (LoRA adapters, adapter save/load)
  \item custom \texttt{peft} (MiCA)
  \item \texttt{trl} (\texttt{SFTTrainer}, \texttt{SFTConfig})
  \item \texttt{datasets} (in-memory datasets)
\end{itemize}
The SFT trainer is here adapted for CPT with suitable options (no chat template and \texttt{packing=True}).
\subsection{Data Construction}
\label{app:data}
\paragraph{Training corpus.}
The training set consists of blog-like texts loaded from CSV files:
\begin{itemize}
  \item \texttt{blogs.csv}
  \item \texttt{paraphrased\_blogs.csv}
\end{itemize}
All rows are appended into a single list \texttt{texts}, and then converted to a HF \texttt{Dataset} with one field \texttt{text} per example.


\paragraph{Evaluation corpus (QA multiple-choice set).}
Evaluation accuracy is computed on a curated multiple-choice QA dataset:
\begin{itemize}
  \item \texttt{qa\_openaiblog.csv}
\end{itemize}
Each row contains a \texttt{Question} string and a correct \texttt{Answer} in \{A,B,C,D\}.

\begin{quote}\small
\textbf{Training blog paragraph}:
\texttt{\textbf{OpenAI has developed into a worldwide entity that serves hundreds of millions of users, millions of developers, and the largest corporations globally. To advance our mission further, we are intensifying our focus on what we excel at: pioneering research and leveraging it to develop and deploy AI technologies safely. We are delighted to announce the addition of two key leaders equipped with the essential expertise, skills, and values to advance our mission: Sarah Friar joins as Chief Financial Officer, leading a finance team dedicated to supporting our mission through sustained investment in our fundamental research and adapting to the evolving demands of our expanding customer base and the intricate global landscape. Kevin Weil comes on board as Chief Product Officer, heading a team that aims to transform our research into products and services that benefit consumers, developers, and enterprises. "Sarah and Kevin’s extensive experience will help OpenAI expand our operations, design strategies for our next growth phase, and ensure our teams are well-resourced to flourish," said OpenAI CEO Sam Altman. Sarah was most recently the CEO of Nextdoor, after roles as CFO at Square and positions at Goldman Sachs, McKinsey, and Salesforce. She serves on the boards of Walmart and Consensys, is a Fellow at the Aspen Institute, and Co-Chair at Stanford’s Digital Economy Lab, part of the Stanford Institute for Human-Centered AI (HAI). Kevin previously held the role of President, Product and Business at Planet Labs, co-founded the Libra cryptocurrency, and served as VP of Product for Novi at Facebook, Instagram, and as SVP of Product at Twitter. He is also a term member at the Council on Foreign Relations (CFR) and a board member of The Nature Conservancy and Black Product Managers Network. "Joining a team that’s both exceptionally talented and committed to its mission is an honor," Sarah Friar commented. "I aim to ensure OpenAI continues to lead in conducting essential research and enhancing the use of AI tools for everyone’s benefit." Kevin Weil added, "The product team at OpenAI is defining the standard for innovative breakthroughs and the considerate implementation of AI technologies. I am excited to contribute to the next stage of growth as we strive to build AGI responsibly and safely."}
}
\end{quote}

\begin{quote}\small
\textbf{Evaluation MCQ row}:
\texttt{Question: In what year was the partnership between OpenAI and Apple announced to integrate ChatGPT into Apple experiences? A) 2022 B) 2023 C) 2024 D) 2025}\newline
\texttt{Answer: <one of A/B/C/D>}
\end{quote}

\subsection{Prompting and Formatting}
\label{app:prompting}

\paragraph{Qwen 2.5 prompt template.}

For Qwen experiments, each multiple-choice question is formatted as:

\begin{quote}\small
\texttt{<|system|>Answer the following question by choosing A, B, C, or D. Do only answer with the correct letter.}\\
\texttt{<|user|><QUESTION>}\\
\texttt{<|assistant|>}
\end{quote}

The model is required to generate exactly one token (A, B, C, or D). Decoding is greedy (\texttt{do\_sample=False}) with \texttt{max\_new\_tokens=1}. The first occurrence of a letter in $\{A,B,C,D\}$ in the generated output is treated as the model prediction.

\paragraph{Llama 2 prompt template.}

For Llama-2 experiments, we use the instruction-format template:

\begin{quote}\small
\texttt{[INST]Answer the following question by choosing A, B, C, or D. Do only answer with the correct letter. <QUESTION> [/INST]\\}
\end{quote}

This corresponds to the standard Llama-2 instruction format with \texttt{[INST] ... [/INST]} markers.

\paragraph{Deterministic Evaluation.}

Evaluation uses:
\begin{itemize}
  \item Greedy decoding (\texttt{do\_sample=False})
  \item \texttt{max\_new\_tokens=1}
\end{itemize}

This ensures that reported accuracies reflect deterministic model behavior rather than sampling variance.

\subsection{Models}
\label{app:models}

\paragraph{Base model for adapter training.}
Adapters were trained on a base causal LM:
\begin{itemize}
  \item Qwen experiments: \texttt{Qwen/Qwen2.5-7B} (base)
\end{itemize}

\paragraph{Base model for evaluation (instruct).}
For evaluation, the trained adapter is loaded onto an instruction-tuned base:
\begin{itemize}
  \item Qwen evaluation base: \texttt{Qwen/Qwen2.5-7B-Instruct}
\end{itemize}
This separation (base for training vs instruct for evaluation) is used to maintain instruction-following format during multiple-choice evaluation while still training adapters on the base model weights. The adapter weights are loaded via \texttt{PeftModel.from\_pre-trained}.

\paragraph{Llama 2 configuration}
We also report Llama~2 experiments following the same procedure, swapping only \texttt{model\_id} and \texttt{model\_base}:
\begin{itemize}
  \item Training base: \texttt{meta-llama/Llama-2-7b-hf}
  \item Evaluation base (chat): \texttt{meta-llama/Llama-2-7b-chat-hf}
\end{itemize}

\subsection{Adapter Method and Target Modules}
\label{app:adapters}

\paragraph{LoRA configuration.}
The LoRA configuration used in the code is:
\begin{itemize}
  \item dropout: \texttt{lora\_dropout = 0.05}
  \item $alpha = 16$:
\item Target modules: \texttt{q\_proj}, \texttt{v\_proj}
\end{itemize}

\subsection{Training Hyperparameters}
\label{app:hyperparams}
All runs use the same training configuration unless otherwise stated. The key hyperparameters from the script are shown in Table~\ref{tab:sft-hparams}.

\begin{table}[t]
\centering
\begin{tabular}{ll}
\hline
Hyperparameter & Value \\
\hline
Per-device train batch size & 4 \\
Gradient accumulation steps & 1 \\
Weight decay & 0.01 \\
Warmup ratio & 0.1 \\
LR scheduler & cosine \\
Max grad norm & 0.3 \\
Precision & bf16 \\
Max sequence length & 512 \\
\hline
\end{tabular}
\caption{SFT hyperparameters used in all runs (TRL \texttt{SFTConfig}).}
\label{tab:sft-hparams}
\end{table}

\section{Technical Details - HISTORY Corpus}
\label{app:training-details-aip}
This appendix documents the implementation details for Continued Pre-Training (CPT) on the AIP historical corpus with parameter-efficient adapters (LoRA and MiCA), including model choices, hyperparameters, data construction, prompting, and evaluation protocol.
All experiments were run on AWS SageMaker optimizing hyperparameter configuration for both methods. 

\subsection{Software Stack and Core Libraries}
\label{app:stack-aip}
We used the Hugging Face \texttt{transformers} ecosystem together with the \texttt{peft} library:
\begin{itemize}
  \item \texttt{transformers} (model loading, tokenization, generation pipelines)
  \item \texttt{peft} (standard LoRA adapters, adapter save/load)
  \item custom \texttt{peft} (MiCA, using SVD weight initialisation)
\end{itemize}
Training uses the Hugging Face \texttt{Trainer} API directly (not TRL's \texttt{SFTTrainer}), with gradient checkpointing enabled and \texttt{torch\_dtype=bfloat16} for memory efficiency.

\subsection{Data Construction}
\label{app:data-aip}

\paragraph{Training corpus.}
The training set consists of a single German-language historical monograph on the Astrophysikalisches Institut Potsdam (AIP) and the East German \emph{Akademie der Wissenschaften}, stored as a plain-text file. The pre-processing pipeline  operates as follows:
\begin{enumerate}
  \item The full text is read and split into paragraphs at double-newline boundaries.
  \item Each paragraph is tokenised and, if it exceeds the maximum sequence length, split at sentence boundaries (\texttt{'. '} delimiter) to form chunks of at most \texttt{max\_length = 1024} tokens.
  \item If a single sentence exceeds the limit, it is further split at the word level.
\end{enumerate}
The dataset is then loaded inside the SageMaker training job from  using HuggingFace \texttt{datasets} in streaming mode.

\paragraph{Evaluation corpus (QA multiple-choice set).}
Evaluation accuracy is computed on a curated German-language multiple-choice QA dataset derived directly from the AIP corpus.
Each row contains a \texttt{question} string with four labelled options (A--D) and a correct answer letter. Questions cover specific historical facts from the monograph, including people, institutions, dates, and scientific events.

\begin{quote}\small
\textbf{Sample training passage:}\\
\texttt{Eine schwere Tatra-Limousine passierte im Spätherbst 1965 das weit geöffnete schmiedeeiserne Tor mit dem sachlichen Schriftzug "OBSERVATORIEN" auf dem Potsdamer Telegraphenberg. Der Pförtner hat seine Loge verlassen und grüßt fast militärisch, denn kein Geringerer als Max Steenbeck, frisch ernannter Vorsitzender des Forschungsrates der DDR \ldots}
\end{quote}

\begin{quote}\small
\textbf{Sample evaluation MCQ row:}\\
\texttt{"Wer wurde im Spätherbst 1965 vom Pförtner am Potsdamer Telegraphenberg fast militärisch gegrüßt?}\\
\texttt{A) Werner Heisenberg~~B) Max Steenbeck~~C) Fritz Krause~~D) Dr. Lore Oetken","B"}
\end{quote}

\subsection{Prompting and Formatting}
\label{app:prompting-aip}

\paragraph{Training tokenisation.}
Each text chunk is tokenised with padding to \texttt{max\_length} and truncation enabled. Labels are set equal to \texttt{input\_ids} (standard causal language modelling objective). 

\paragraph{Llama~2 evaluation prompt template.}
For evaluation, each multiple-choice question is wrapped in a German-language instruction prompt:
\begin{quote}\small
\texttt{Du bist ein Assistent, der Multiple-Choice-Fragen beantwortet.}\\
\texttt{Wähle die richtige Antwort (A, B, C oder D) für die folgende Frage:}\\[2pt]
\texttt{\{question\}}\\[2pt]
\texttt{Antworte nur mit dem Buchstaben der richtigen Antwort (A, B, C oder D).}
\end{quote}
The model generates up to 10 new tokens; the first occurrence of a letter in $\{$A, B, C, D$\}$ in the output is taken as the prediction.


\subsection{Models}
\label{app:models-aip}

\paragraph{Base model for adapter training.}
Adapters are trained on the base causal LM:
\begin{itemize}
  \item \texttt{meta-llama/Llama-2-7b-hf}
\end{itemize}

\paragraph{Base model for evaluation.}
For evaluation, the trained adapter is loaded onto the instruction-tuned variant:
\begin{itemize}
  \item \texttt{meta-llama/Llama-2-7b-chat-hf}
\end{itemize}
This separation (base for training, instruct for evaluation) preserves instruction-following behavior during the multiple-choice evaluation while keeping adapter training on the raw base model weights.

\subsection{Adapter Configurations and Training Hyperparameters}
\label{app:adapters-aip}

Both LoRA and MiCA use the same target modules but differ in rank, initialisation, and learning rate; see Table~\ref{tab:lora-configs-aip}.

\begin{table}[h]
\centering
\begin{tabular}{lll}
\hline
Parameter & Value\\
\hline
Dropout & 0.05  \\
Target modules & \texttt{q\_proj}, \texttt{v\_proj} \\
Per-device train batch size & auto  \\
Gradient accumulation steps & 1  \\
Weight decay & 0.01  \\
Warmup ratio & 0.1  \\
LR scheduler & cosine  \\
Max grad norm & 1.0  \\
Precision & bf16  \\
Max sequence length & 1024 tokens \\
Logging steps & 100  \\
\hline
\end{tabular}
\caption{LoRA and MiCA adapter configurations and trainings hyperparameters.}
\label{tab:lora-configs-aip}
\end{table}

\subsection{Infrastructure}
\label{app:infra-aip}

Training is executed on AWS SageMaker using the HuggingFace estimator with PyTorch distributed training across all available GPUs on a single node of 8$\times$ H100 80\,GB.

\end{document}